\documentclass{article}
\usepackage{ijcai20}

\usepackage{times}
\usepackage{soul}
\usepackage{url}
\usepackage{xcolor}
\usepackage[hidelinks]{hyperref}
\usepackage[utf8]{inputenc}
\usepackage[small]{caption}
\usepackage{graphicx}
\usepackage{amsmath}
\usepackage{amsthm}
\usepackage{amssymb}
\usepackage{booktabs}
\usepackage{algorithm}
\usepackage{algorithmic}
\usepackage{adjustbox}
\usepackage{caption} 
\usepackage{multirow}
\usepackage{scrextend}
\usepackage{float}
\usepackage{footmisc}

\title{Handling Variable-Dimensional Time Series with Graph Neural Networks \footnote{Accepted at AI4IoT@IJCAI'20 workshop}}
\author{
Vibhor Gupta\footnote{Equal Contribution}
\and
Jyoti Narwariya$^\dagger$\and
Pankaj Malhotra\and
Lovekesh Vig\and
Gautam Shroff 
\affiliations
TCS Research, New Delhi, India
\emails
\{g.vibhor, jyoti.narwariya, malhotra.pankaj, lovekesh.vig, gautam.shroff\}@tcs.com
}

\begin{document}
\maketitle
\begin{abstract}
Several applications of Internet of Things (IoT) technology involve capturing data from multiple sensors resulting in multi-sensor time series. 
Existing neural networks based approaches for such multi-sensor or multivariate time series modeling assume fixed input dimension or number of sensors.
Such approaches can struggle in the practical setting where different instances of the same device or equipment such as mobiles, wearables, engines, etc. come with different combinations of installed sensors.
We consider training neural network models from such multi-sensor time series, where the time series have varying input dimensionality owing to availability or installation of a different subset of sensors at each source of time series. 
We propose a novel neural network architecture suitable for zero-shot transfer learning allowing robust inference for multivariate time series with previously unseen combination of available dimensions or sensors at test time.
Such a combinatorial generalization is achieved by conditioning the layers of a core neural network-based time series model with a ``conditioning vector'' that carries information of the available combination of sensors for each time series. 
This conditioning vector is obtained by summarizing the set of learned ``sensor embedding vectors'' corresponding to the available sensors in a time series via a graph neural network.
We evaluate the proposed approach on publicly available activity recognition and equipment prognostics datasets, and show that the proposed approach allows for better generalization in comparison to a deep gated recurrent neural network baseline.
\end{abstract}

\section{Introduction}
Multi-sensor time series data is ubiquitous and growing at a rapid pace owing to the increasing ambit of devices under the Internet of Things \cite{atzori2010internet,da2014internet} technology.
Deep learning approaches have been successfully explored for applications in multivariate time series forecasting, classification, anomaly detection, and remaining useful life estimation \cite{p:lstm-ad,lai2018modeling,ordonez2016deep,heimes2008recurrent}.

However, most existing approaches for multivariate time series data assume a fixed-dimensional time series as input.
In many practical settings, such an assumption may not hold. 
For instance, when learning activity recognition models, time series from different people may involve a varying number of available sensors (such as accelerometer, gyroscope, and magnetometer) owing to different wearable  or mobile devices. 
Similarly, equipment health monitoring models have to deal with data from different equipment instances with varying set of sensors (such as temperature, vibration, and pressure) installed on them.

In this work, we consider the setting where multiple multivariate time series are generated from different instances of the same underlying dynamical system (e.g. humans in activity recognition, or engines in equipment health monitoring), with different instances having a different combination of available sensors.
A simple approach to deal with a missing sensor in a time series instance is to assume a constant (e.g. mean) value for that sensor based on the statistics from other time series instances where the sensor is available \cite{che2018recurrent}.
We empirically show that the performance of such an approach degrades rapidly as the percentage of missing sensors in the test instance increases.
Furthermore, many approaches dealing with missing values in a time series, use data imputation methods such as smoothing, interpolation, and spline, etc. \cite{che2018recurrent}, which are not directly applicable in this setting as they rely on the availability of at least one value for each dimension in the time series.

Another potential approach is to train a different network for each possible combination of available sensors, which 1) is not scalable as the number of possible combinations grows exponentially, 2) assumes availability of sufficient training data for each combination, and 3) does not retain any knowledge across combinations.
Instead, we propose leveraging a neural network architecture with two modules where a core module caters to modeling the temporal aspect of the data while another conditioning module caters to adjusting the core module based on the combination of sensors available in each time series, effectively exhibiting different behavior depending on the available sensors. 
Based on empirical evaluation on two activity recognition datasets and a prognostics dataset, we show that the proposed approach outperforms the baseline approach which uses mean-imputation for the missing sensors in two settings: 
i. \textit{zero-shot setting} where the combination of available sensors at test time is different from any of the combinations in the training set, 
ii. \textit{fine-tuning setting}: where along with the fixed training set with different combinations, a small number of instances for the same combination as that of the test instance are available for fine-tuning.

\section{Related Work}
The literature on the problem of handling variable-dimensional input in multivariate time series is scarce.
Though several approaches in the neural networks literature deal with the problem of varying dimensionality, most of these are primarily restricted to studying variability along the time dimension or the pixel space in images and videos, where these are naturally handled, e.g. via recurrent neural networks \cite{hochreiter1997long} for the time dimension, and variants of pooling operations for the image applications \cite{he2015spatial}. 

Recently, a neuro-evolutionary approach has been proposed in \cite{elsaid2020neuro} which studies the problem of structure-adaptive transfer learning for time-series prediction. It relies on a series of mutation operations and crossover (reproduction) operations over the neural units. Instead of the computationally expensive neuro-evolutionary approaches, we take a different perspective on the problem of adapting to varying dimensionality where graph neural networks (GNNs) are used to achieve combinatorial generalization. Such generalization ability in GNNs has been recently studied in different contexts, e.g. for learning structured reinforcement learning (RL) policies where the same nodes in the GNN can be re-used as modules to learn RL agents with different overall graph structure. For instance, NerveNet \cite{wang2018nervenet} shows the ability of GNNs to transfer knowledge from a four-legged robot to a six-legged robot.

Recent work on Modular Universal Reparameterization (MUiR) \cite{meyerson2019modular} attempts to learn core neural modules which can be transferred across varying input and output dimensions. It proposes learning the core module by solving several sets of architecture-task problems with varying input and output dimensionality. However, MUiR does not study multivariate time series setting and relies on solving several tasks to learn the core neural module. Instead, our approach relies on a conditioning vector obtained via GNNs to allow adaptability to varying input dimensions.

The recently proposed CondConv \cite{yang2019condconv} is similar in spirit to our work, i.e. it attempts to dynamically adapt the parameters of the neural network conditioned on the current input. While it focuses on adapting to each input with same dimensionality, our work focuses on adapting to inputs of varying dimensionality. Though significantly different in implementation and the end-objective, our approach draws inspiration from such works, including \cite{rosenbaum2019routing,andreas2016neural}, where the parameters of the core neural network are dynamically adjusted as per the input.

Handling variable input dimensionality can be seen as a harder case for the missing value problem in time series.
Several approaches for handling missing values in multivariate time series via neural networks have been proposed. 
For instance, \cite{che2018recurrent} study missing value problem in multivariate time series by proposing a variant of the gated recurrent units \cite{cho2014learning} using knowledge of which dimensions of the input are missing and for how long. Such approaches are, however, not directly applicable in our setting where one or more dimensions of the time series are completely missing, i.e. the missing percentage is 100\%, since they rely on one or more past values to adapt.

Several transfer learning and gradient-based meta learning approaches \cite{timenet,fawaz2018transfer,kashi2019convtimenet,narwariya2020meta} have been recently proposed for time-series classification. However, they study the problem of quickly adapting to new classes of interest or new domains in the univariate setting, and do not address the problem of varying-dimensional multivariate time series. 
In future, our work can potentially be used to extend the transfer and few-shot learning capabilities of these approaches to varying-dimensional time series across domains.

More specific to the activity recognition task that we study as an application of our approach, \cite{wang2018stratified} proposes a transfer learning approach called Stratified Transfer Learning (STL) for cross-domain activity recognition. The source and target domains are the body parts of the same person or different person, e.g., transfer knowledge from left arm to right arm, right arm to the torso, etc. This approach considers knowledge transfer based on the similarity of installed sensors on body parts. On the other hand, our approach considers transferring knowledge to different combinations of sensors and is therefore, orthogonal in nature to STL.
Heterogeneous transfer learning  has been studied in activity recognition \cite{feuz2015transfer}, where the idea is to learn a mapping of sensors from the source domain to those in the target domain when the meta-data related to sensors is unknown restricting easy mapping of dimensions across datasets.
This setting is orthogonal to our setting and either approaches can potentially benefit from each other.

\begin{figure*}
	\centering
	\includegraphics[width=0.7\textwidth,trim={5cm 28.5cm 20cm 3cm}, clip]{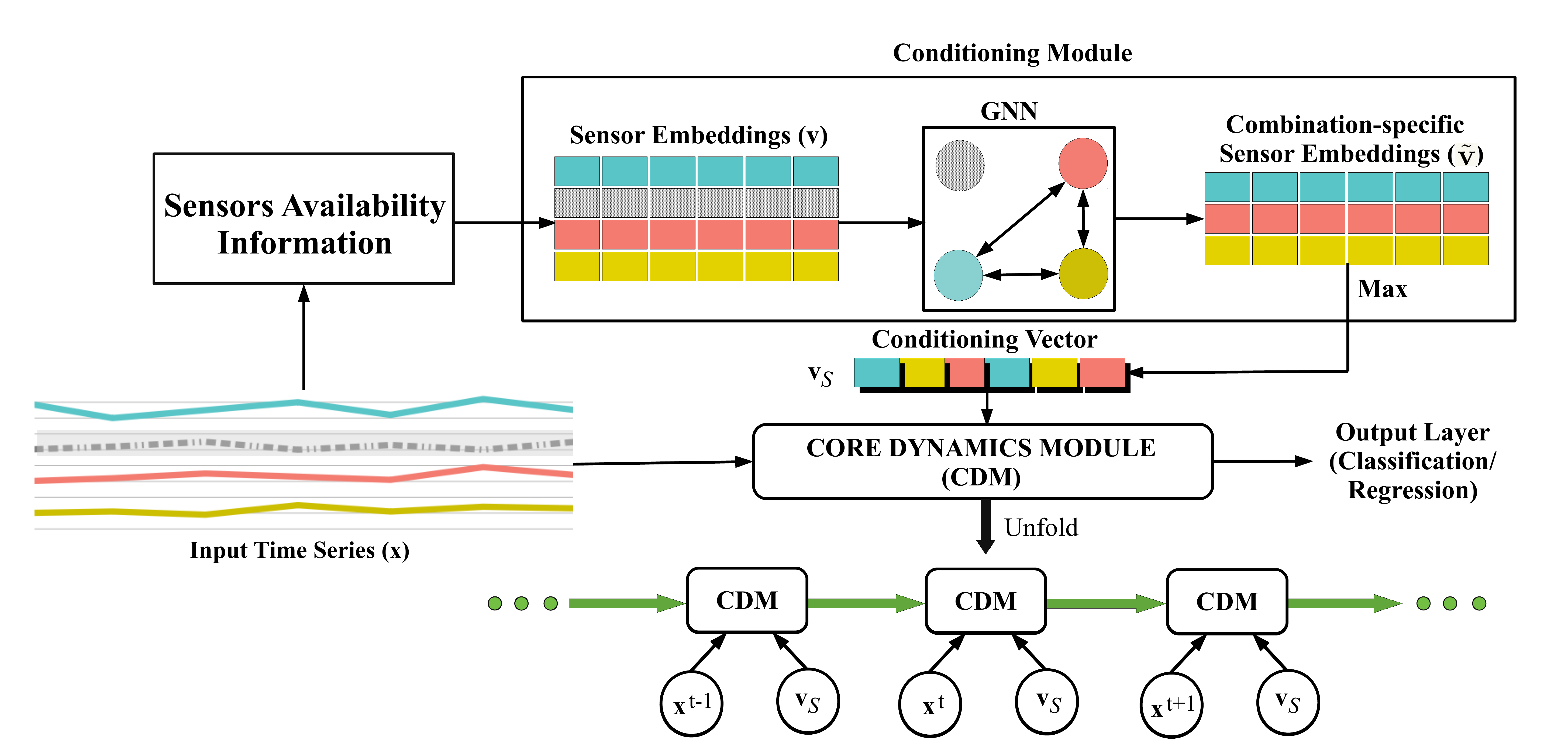}
	\caption{Illustrative flow-diagram for the proposed approach. The available sensors in a time series and corresponding active embeddings and nodes are shown in color while the missing sensor/dimension is shown in grey color (dashed). The active nodes corresponding to the available sensors exchange messages with each other via a graph neural network, and generate a conditioning vector that is used by the core dynamics module (a recurrent neural network) processing the time series to estimate the target. \label{fig:flowchart}}
\end{figure*}

\section{Problem Definition}

Consider a training set $\mathcal{D} = \{(\mathbf{x}_i,y_i)\}_{i=1}^N$ having $N$ multivariate time series $\mathbf{x}_i \in \mathcal{X}$ with an associated target $y_i \in \mathcal{Y}$.
Each time series $\mathbf{x}_i = \{\mathbf{x}_i^t\}_{t=1}^{T_i}$ has length $T_i$ with $\mathbf{x}_i^t \in \mathbb{R}^{d_i}$, where $1\leq d_i\leq d$ is the number of available dimensions or sensors out of a total of $d$ sensors. Further, let $\mathcal{S}$ denote the set of all the sensors. Then, the $d_i$ dimensions for the $i$-th time series correspond to a subset of sensors $ \mathcal{S}_i \subseteq \mathcal{S}$, with $|\mathcal{S}_i|=d_i$ and $|\mathcal{S}|=d$.
The goal is to learn a mapping $f\!\!:\mathcal{X} \!\rightarrow \mathcal{Y}$.
Note that this setting is different from the standard multivariate time series modeling setting in the sense that  the input dimension $d_i$ is allowed to vary across the time series in $\mathcal{D}$.
When $d$ is reasonably large\footnote{In the three datasets we consider, $d$ is 9, 21, and 45.}, the subsets of $\mathcal{S}$ available across time series in $\mathcal{D}$ tends to be much smaller than the set of all possible combinations, and therefore, good generalization at test time requires learning a function that provides robust estimates beyond the combinations of sensors seen in $\mathcal{D}$.

Though easily extendable to other applications of multivariate time series modeling, we consider two tasks in this work: i. classification (for activity recognition), and ii. regression (for remaining useful life estimation). 
For the $K$-way classification tasks, target $y_i$ corresponds to one of the $K$ classes $\{1,\ldots,K\}$. 
When using neural networks as approximators for $f$, $y_i\in\{0,1\}^K$ is represented as a one-hot vector, with value being 1 for the dimension corresponding to the target class, and 0 for the remaining $K-1$ dimensions.
For activity recognition, the $K$ classes correspond to $K$ different activities.
For the regression task, $y_i \in \mathbb{R}$; in case of remaining useful life (RUL) estimation, $y_i = F_i - T_i$, where $F_i$ denotes the total operational life of an equipment instance $i$ till it fails, with $1 \leq T_i \leq F_i$. 

\section{Approach\label{sec:approach}}
We represent each point $\mathbf{x}_i^t$ in $\mathbb{R}^d$ irrespective of the number of available sensors $d_i$, where the unavailable sensors are mean-imputed, i.e. the mean value of the sensor across other instances where it is available is used. This allows us to input a fixed-dimensional time series to the core module that aims to capture the temporal patterns or dynamics in the data. Along with this core module, we provision for an additional conditioning module which takes into account the sensors available in a particular time series, and generates a conditioning vector that accordingly alters the processing of the core module. 

More specifically, our approach consists of the following two modules, as illustrated in Figure \ref{fig:flowchart}:
\begin{enumerate}
    \item a \textit{core dynamics module} which is a (gated) recurrent neural network \cite{cho2014learning} that learns the dynamics of the system and ingests a fixed-dimensional multivariate time series where the missing dimensions are imputed with a constant (mean) value, 
    \item a \textit{conditioning module} which generates a ``conditioning vector'' as a function of the available sensors irrespective of the readings those sensors take for a particular time series instance. 
    This conditioning vector is passed as an additional input to the core dynamics module allowing it to adjust its internal computations and activations according to the combination of available sensors.
    The conditioning vector is in-turn obtained from ``sensor embedding vectors'' via a graph neural network \cite{scarselli2008graph}.
\end{enumerate}
Note that the core dynamics module and the conditioning module along with the sensor embeddings are jointly learned in an end-to-end fashion via stochastic gradient descent (SGD).

Each sensor is associated with a vector or embedding, and the vectors for any given combination of available sensors in a time series are used to obtain the combination-specific conditioning vector.
This is equivalent to mapping a set of (sensor) vectors to another (conditioning) vector. 
Though the core module based on RNNs can only ingest fixed-dimensional time series input, the conditioning vector can be obtained by summarizing a variable number of sensor vectors via a GNN \cite{scarselli2008graph,battaglia2018relational}, as detailed in the next sub-section. This conditioning vector serves as an additional input that allows the core module to adjust its processing according to the variable number of available sensors within each time series.
A key advantage of using a GNN for processing the combination of sensors is that once the GNN is learned, it can process any previously unseen combination of sensors apart from those seen during training, thus depicting combinatorial generalization \cite{battaglia2018relational}. 
We empirically show (later in Section \ref{sec:exp}) that this is indeed the case, and the proposed conditioning module does allow the core module to robustly adapt to previously unseen sensor combinations at test time.
We first explain how we obtain the conditioning vector using a GNN-based conditioning module, and then explain the RNN-based core dynamics module. 

\subsection{Conditioning Module}
Each sensor $s \in \mathcal{S}$ is associated with a learnable embedding vector $\mathbf{v}_s \in \mathbb{R}^{d_s}$. 
Corresponding to the set of sensors $\mathcal{S}$, consider a graph $\mathcal{G}(\mathcal{V},\mathcal{E})$ with nodes or vertices $\mathcal{V}$ and edges $\mathcal{E}$, with one node $v_s \in \mathcal{V}$ for every $s \in \mathcal{S}$ such that $|\mathcal{V}|=|\mathcal{S}|$.
The neighbors of any node $v_s$ are denoted by $\mathcal{N}_{\mathcal{G}}(v_s)$.

For a specific combination $\mathcal{S}_i \subseteq \mathcal{S}$, only the nodes $\mathcal{V}_i \subseteq \mathcal{V}$ corresponding to the sensors in $\mathcal{S}_i$ are considered to be \textit{active}, and contribute to obtaining the combination-specific conditioning vector $\mathbf{v}_{\mathcal{S}_i}$. 
For the active nodes, the graph is assumed to be \textit{fully-connected} such that each active node in the graph is connected to every other active node. 
As depicted in Figure \ref{fig:flowchart}, any edge is active only if both the participating nodes are active.

The GNN corresponding to this graph consists of a node-specific feed-forward network $f_n$ and an edge-specific feed-forward network $f_e$; $f_n$ and $f_e$ are shared across the nodes and edges in the graph, respectively.
For any active node $v_k \in \mathcal{V}_i$, the node vector $\mathbf{v}_k$ is updated using the GNN as follows: 
\begin{align}
\mathbf{v}_{kl} &= f_{e}([\mathbf{v}_{k},\mathbf{v}_{l}]; \boldsymbol{\theta}_{e}) \label{eq1}, \quad \forall v_l \in \mathcal{N}_{\mathcal{G}}(v_k),\\
\tilde{\mathbf{v}}_{k} &= f_{n}([\mathbf{v}_{k},\sum_{\forall l} \mathbf{v}_{kl}]; \boldsymbol{\theta}_{n}) \label{eq2},
\end{align}
where $f_{e}$ and $f_{n}$ both consist of leaky ReLU \cite{maas2013rectifier} layers with learnable parameters $ \boldsymbol{\theta}_{e}$ and $ \boldsymbol{\theta}_{n}$, respectively. We also use dropout after these layers for regularization \cite{srivastava2014dropout}. 
While $f_{e}$ computes the message from node $v_l$ to $v_k$, $f_{n}$ updates the node vector $\mathbf{v}_k$ to  $\tilde{\mathbf{v}}_k$ using the aggregated message from its neighboring nodes. 
Finally, the conditioning vector $\mathbf{v}_{\mathcal{S}_i}\in \mathbb{R}^{d_s}$ specific to a combination of sensors is obtained from the updated node vectors as
\begin{equation}\label{eq3}
    \mathbf{v}_{\mathcal{S}_i} = \mathtt{max}(\{\tilde{\mathbf{v}}_k\}_{v_k \in \mathcal{V}_i}), 
\end{equation}
where $\mathtt{max}$ returns the dimension-wise maximum value\footnote{We also tried averaging instead of max operation to summarize the vectors of available sensors, but found max to work better on all datasets considered.} across the updated node vectors for the particular combination of sensors $\mathcal{S}_i$.
It is noteworthy that the summation over the messages across nodes in Equation \ref{eq2} and the $\mathtt{max}$ operation in Equation \ref{eq3} essentially provide the desired ability to process varying number of nodes (sensors) in the conditioning module. 

\subsection{Core Dynamics Module}
As mentioned earlier, any time series $\mathbf{x}_i \in \mathcal{D}$ is first converted to the $d$-dimensional time series $\tilde{\mathbf{x}}_i$ with mean-imputation for the unavailable sensors.
This time series along with its conditioning vector $\mathbf{v}_{\mathcal{S}_i}$ are processed by the core dynamics model as follows:
\begin{align}
\mathbf{z}^t_i &= GRU([\tilde{\mathbf{x}}_i^t, \mathbf{v}_{\mathcal{S}_i}],\mathbf{z}^{t-1}_i; \boldsymbol{\theta}_{GRU}), \quad t:1,\ldots,T_i\\
\hat{y}_i &=f_{o}(\mathbf{z}^{T_i}_i;\boldsymbol{\theta}_{o})\label{eq5},
\end{align}
where $GRU$ is a (multi-layered) GRU-based RNN \cite{cho2014learning} having $\boldsymbol{\theta}_{GRU}$ learnable parameters that gives feature vector $\mathbf{z}^{T_i}$ at the last time step $T_i$. 
At last, the estimate $\hat{y}_i$ for $y_i$ is obtained via $f_{o}$ consisting of ReLU layer(s) followed by softmax or sigmoid layer depending upon whether the task is classification or regression, respectively, such that $\hat{y}_i \in [0,1]^K$ in case of classification, and $\hat{y}_i \in \mathbb{R}$ in case of regression. For the RUL estimation regression task, we use min-max normalized targets such that they lie in $[0,1]$.

\subsection{Training objectives}
We use the standard cross-entropy $\mathcal{L}_c$ and squared-error $\mathcal{L}_r$ losses as training objectives for the classification and regression tasks, respectively: 
\begin{align}
\mathcal{L}_{c}=-\frac{1}{N}\sum_{i=1}^{N}\sum_{k=1}^{K}{y_i^k log(\hat{y}_i^k}) \label{eq:cl},\\
\mathcal{L}_{r} = \frac{1}{N}\sum_{i=1}^{N} (y_i-\hat{y}_i)^2 \label{eq:rl},
\end{align}
where $y_i^{k}$ denotes the $k$-th dimension of $y_i$, etc.
All the parameters $\boldsymbol{\theta}_{n}$, $\boldsymbol{\theta}_{e}$, the vectors $\mathbf{v}_s$ for all $s\in \mathcal{S}$, $\boldsymbol{\theta}_{GRU}$, and $\boldsymbol{\theta}_{o}$ are learned via mini-batch SGD.
In practice, we consider time-series with the same set of available sensors within a mini-batch such that the active nodes in $\mathcal{G}$ for all the time series in a mini-batch are the same.

\section{Experimental Evaluation\label{sec:exp}}

\subsection{Datasets Description}
\begin{table}[h!]
\centering
\scalebox{0.8}{
\begin{tabular}{|l|c|c|c|c|c|}
\hline
Dataset & $d$ & Task & \textbf{$N$} & $K$ \\
\hline
DSADS & 45  & C & 9,120 & 19 \\
HAR & 9  & C & 10,299 & 6    \\
Turbofan & 21  & R & 519 & - \\
\hline
\end{tabular}}
\caption{Dataset details. Here, $d$: maximum available sensors in a time series, C: classification, R: regression, $N$: number of instances, $K$: number of classes.}
\label{tab:datasets}
\end{table}
As summarized in Table \ref{tab:datasets}, we use two publicly available activity recognition benchmark datasets used by \cite{karim2019multivariate}\footnote{\url{https://github.com/titu1994/MLSTM-FCN/releases}}) and an RUL estimation Turbofan Engine dataset: 

\textbf{DSADS} (Daily and Sports Activities Data Set) \cite{altun2010human}: DSADS contains 45 sensors and 19 activities such that each activity is recorded using sensors installed in different body parts, e.g., torso, right arm, left arm, right leg, and left leg. 
\textbf{HAR} (Human Activity Recognition Using Smartphones) \cite{anguita2012human}: HAR contains 9 sensors and 6 activities (walking, walking\_upstairs, walking\_downstairs, sitting, standing, laying) using a smartphone on the waist. 
\textbf{Turbofan Engine\footnote{\url{https://ti.arc.nasa.gov/tech/dash/groups/pcoe/prognostic-data-repository/##turbofan}} (FD002)} \cite{saxena2008damage}: We use the FD002 dataset of the turbofan engines dataset repository containing time series of readings for 21 sensors and 3 operating condition variables, such that each cycle in the life of an engine provides a 24-dimensional vector.
The sensor readings for the engines in the training set are available from the starting of their operational life till the end of life or failure point, while those in the test set are clipped at a random time prior to the failure, and the goal is to estimate the RUL for these test engines. 

\subsection{Experimental Setup}
We evaluate our proposed approach in two settings: 
i. \textit{Zero-shot}, and ii. \textit{Fine-tuning}.
In either setting, sensor combinations unseen at training time are used while testing, therefore, evaluating the limit of the network to generalize to previously unseen sensor combinations. 
In zero-shot setting, the trained network is directly used for inference, whereas in fine-tuning setting, a small fraction of labeled time series with the same sensor combination as that of the test instance is used for fine-tuning the network.

Let $f_{tr}$ and $f_{te}$ denote the fraction of unavailable sensors for each time series during training and testing, respectively.
We evaluate for $f_{tr}=\{0.25,0.4\}$, and $f_{te}=\{0.1,0.25,0.4,0.5\}$.
We restrict the number of available sensor combinations at training time to 16 in order to mimic a realistic scenario where multiple instances for a particular sensor combination would be available while the number of such combinations itself would be small. From within these 16 combinations, further combinations are generated by randomly masking additional sensors, resulting in a total of 64 sensor combinations seen during training. 

For all datasets, around 40\% of the instances are used for training, 10\% for validation, 10\% fine-tuning (ignored in zero-shot case), and remaining 40\% for testing.
For Turbofan dataset, each time series is divided into overlapping windows of length 100 with a shift of 5.
For DSADS and HAR, we z-normalize each input dimension using sensor-wise mean and standard deviation from the train set, whereas sensor-wise min-max normalization is used for Turbofan dataset.
We use classification error rates and root mean squared error (RMSE) as the performance metrics for the classification and regression tasks, respectively. 

\subsection{Hyperparameters Used}
For all the datasets, the core dynamics module consists of three GRU layers with 128 units each. 
The dimension $d_s$ for the sensor embedding vectors or the node vectors and the resulting conditioning vector is chosen to be $\lfloor\frac{d}{2}\rfloor$.
We use mini-batch size of 64 and 32 for training and 32 fine-tuning, respectively.
All feedforward layers are followed by dropout \cite{srivastava2014dropout} of 0.2 for regularization in addition to early stopping with a maximum of 150 epochs for training and 50 epochs for fine-tuning.
We use vanilla SGD optimizer without momentum to update the sensor embedding vectors with a learning rate of $5e-4$ and Adam optimizer \cite{kingma2015adam} to update the rest of the layers with a learning rate of $1e-4$ . 
Since the active nodes change in every mini-batch with changing combinations of the available sensors, we found it useful to use vanilla SGD for updating the sensor vectors (else, if we use momentum, the vectors for the inactive nodes would also get updated). On the other hand, the GNN and the core dynamics module which are shared across all combinations and mini-batches benefit from momentum, and hence Adam is used for updating their parameters.

\subsection{Baselines Considered}
We refer to our approach as \textbf{GRU-CM} (GRU with GNN-based conditioning module). For comparison, we consider the following approaches:
\begin{enumerate}
\item \textbf{GRU}: This is the baseline approach where the dimensions corresponding to the missing sensors are filled with mean value for that sensor. This is equivalent to the proposed approach but without the conditioning module such that GRU is not provided with any additional signal (the conditioning vector) to allow it to adapt to variable dimensionality of the input.
\item \textbf{GRU with All Sensors Available (GRU-A)}: For all training and testing instances, all sensors are assumed to be available and a GRU network with same hyperparameters as used in our approach is trained. This provides an upper bound for evaluation. 
\item \textbf{GRU with Maxpool over Sensor Embeddings (GRU-SE)}: This is an ablation study over GRU-CM where the steps involved in Equations \ref{eq1}-\ref{eq2} are ignored, and the max operation in Equation \ref{eq3} is directly applied to the original sensor embedding vectors without any combination-specific processing via GNNs. In other words, the active nodes for a particular combination do not exchange messages with each other to allow adaptation to the particular sensor combination at hand. Instead, the embeddings of the active nodes are directly summarized via the max operation in Equation \ref{eq3}.
\end{enumerate}

Note that another baseline could be to learn a separate model for each dimension, but that can be computationally expensive as the resources required would grow with the dimensionality of the time series. Furthermore, such a baseline would not be able to capture interdimensional correlations efficiently.

As a sanity-check, we also trained the GRU baseline (without any conditioning modules) from scratch using just the (10\%) fine-tuning data available for the test-time sensors combination. This results in error rates $\geq 30\%$ for DSADS and HAR across all \%age missing tests (with results degrading with increasing missing \%age), and RMSE of around 60.0 for Turbofan dataset, showing the necessity of using the time series instances from other sensor combinations present in the train set.

\subsection{Results and Observations}
\begin{table*}[ht]
\centering
\caption{Results comparing the proposed approach GRU-CM with other baselines. Classification error rates are reported for DSADS and HAR, while RMSE is reported for Turbofan. (Lower numbers are better). GRU-A is the upper bound assuming all the sensors are available at train and test time. The numbers in bold are statistically significant from the rest with $p < 0.01$.}
\scalebox{0.7}{
\begin{tabular}{|p{1.2cm}|c||ccc|ccc||ccc|ccc||c|}
\hline
\multirow{3}{*}{Dataset} & \multirow{3}{*}{\begin{tabular}[c]{@{}c@{}}$f_{te}$\\  \end{tabular}}  & \multicolumn{6}{c||}{$f_{tr}=0.25$} & \multicolumn{6}{c||}{$f_{tr}=0.4$}& \multirow{2}{*}{\begin{tabular}[c]{@{}c@{}}$f_{tr}=0$,\\$f_{te}=0$  \end{tabular}} \\ \cline{3-14}
&  &  \multicolumn{3}{c|}{Zero Shot} & \multicolumn{3}{c||}{Fine-tune} & \multicolumn{3}{c|}{Zero Shot} & \multicolumn{3}{c||}{Fine-tune} &\\ \cline{3-15} 
 &  &  GRU & {GRU-SE} & {GRU-CM} & {GRU} & {GRU-SE} & {GRU-CM} & {GRU} & {GRU-SE} & {GRU-CM} & {GRU} & {GRU-SE} & {GRU-CM}&{GRU-A} \\ \hline
 \multirow{4}{*}{DSADS}  & 10 & 2.5 & 2.5 & \textbf{1.8} & 1.8 & 2.3 & \textbf{1.7} & 4.6 & 3.7 & \textbf{2.8} & 2.5 & 2.4 & \textbf{1.9} &1.5 \\  
  & 25   & 5.5 & 4.9 & \textbf{3.5} & 2.3 & 2.9 & \textbf{2.1} & 4.6 & 3.1 & \textbf{2.9} & 3.3 & 2.6 & \textbf{2.2} &1.5 \\ 
  & 40  & 12.0 & 13.0 & \textbf{8.1} & 4.6 & 4.6 & \textbf{3.7} & 8.3 & 7.8 & \textbf{5.9} & 4.7 & 3.4 & \textbf{3.3}&1.5 \\  
  & 50  & 21.6 & 21.7 & \textbf{15.9} & 6.6 & 8.0 & \textbf{6.4} & 10.4 & 10.8 & \textbf{8.9} & 5.2 & 5.3 & \textbf{4.4}&1.5  \\ \hline
\multirow{4}{*}{HAR}  & 10  & 9.2 & \textbf{7.0} & 7.5 & 8.4 & \textbf{6.5} & 7.1 & 9.1 & 9.3 & \textbf{7.9} & 9.0 & 8.6 & \textbf{7.5}& 6.3  \\  
  & 25  & 10.4 & \textbf{8.3} & 8.7 & 9.3 & 8.1 & \textbf{7.2} & 10.1 & 10.4 & \textbf{9.1} & 9.5 & 9.4 & \textbf{8.5}  & 6.3 \\  
  & 40  & 12.9 & \textbf{9.7} & 11.1 & 11.1 & \textbf{9.7} & 10.0 & 12.8 & 11.8 & \textbf{11.4} & 12.1 & 10.7 & \textbf{10.5} & 6.3  \\  
  & 50  & 16.1 & \textbf{14.6} & 15.4 & 14.7 & 14.0 & \textbf{13.9} & 16.1 & 15.0 & \textbf{13.8} & 14.6 & 14.7 & \textbf{13.3} & 6.3  \\ \hline

\multirow{4}{*}{\begin{tabular}[c]{@{}c@{}}Turbo-\\    fan\end{tabular}} & 10   & 24.2 & 24.9 & \textbf{24.0} & 23.2 & 23.4 & 23.3 & 24.9 & 24.8 & \textbf{24.4} & 24.1 & \textbf{23.3} & 24.0& 22.4 \\ 
 & 25  & 23.9 & 24.5 & \textbf{23.5} & 22.8 & 23.7 & \textbf{22.7} & 24.7 & 24.8 & \textbf{24.4} & 24.2 & 24.3 & \textbf{24.0}& 22.4 \\  
 & 40  & 26.0 & 26.8 & \textbf{25.7} & 24.6 & 26.3 & \textbf{24.4} & 26.4 & 25.3 & \textbf{25.2} & 26.4 & 25.2 & \textbf{24.7}& 22.4 \\  
 & 50  & 27.1 & 26.8 & \textbf{26.2} & 25.4 & 25.5 & \textbf{24.9} & 26.7 & 26.3 & \textbf{25.9} & 26.3 & 25.6 & \textbf{25.2}& 22.4 \\ \hline
 
\end{tabular}
}
\label{tab:results}
\end{table*}

We make the following key observations from the the results in Table \ref{tab:results}:

\begin{itemize}
   \item The proposed GRU-CM consistently outperforms the vanilla GRU on all three datasets across zero-shot and fine-tuning testing scenarios. In other words, GRU-CM is able to bridge the gap between GRU-A and GRU significantly in most cases proving its robustness to adapt to unseen sensor combinations.
   \item GRU-CM shows significant gains over GRU in zero-shot setting. While fine-tuning for the sensor combinations at test time improves the results for GRU as well as GRU-CM, GRU-CM performs better than GRU depicting better ability to adapt with small amounts of fine-tuning data.
   \item As $f_{te}$ increases, i.e. as the fraction of unavailable sensors at test time increases, the performance of both GRU and GRU-CM degrades. However, importantly, the performance of GRU-CM degrades much more gracefully in comparison GRU showing the advantage of the conditioning module. 
    \item In the ablation study comparing GRU-CM with GRU-SE, we observe that in most cases GRU-CM performs better than GRU-SE. Furthermore, GRU-SE sometimes performs worse than the vanilla GRU. These observations prove the importance of message passing amongst the available sensors to provide better conditioning vectors.
   \item Though sensor combinations used for testing are strictly different from those seen during training, we consider fine-tuning the trained GRU/GRU-CM models using the existing training instances which have the highest overlap with the sensor combination in the test instance. So, instead of relying on new data for fine-tuning the models as studied in Table \ref{tab:results}, we use the closest matching data from the training set in terms of sensor overlap, and conduct a side-study on DSADS. As shown in Table \ref{tab:overlapping}, we observe that the results for both vanilla GRU and GRU-CM improve in comparison to zero-shot method, but GRU-CM still performs better than the fine-tuned vanilla GRU method. This highlights the ability of GRU-CM to adapt to new sensor combinations even better than the costly instance-specific fine-tuning of vanilla GRU.
    
\end{itemize}
\begin{minipage}[t]{0.5\textwidth}
\begin{center}

\begin{adjustbox}{center, width=200pt}

\begin{tabular}{|c||cc|cc|}
\hline
\multicolumn{5}{|c|}{$f_{tr}=0.25$}           \\ \hline
\multirow{3}{*}{\begin{tabular}[c]{@{}c@{}} $f_{te}$\end{tabular}} & \multicolumn{2}{c|}{Zero-Shot} & \multicolumn{2}{c|}{Fine-tune} \\ \cline{2-5} 
                                      & GRU & GRU-CM  & GRU & GRU-CM  \\ \hline 
\multirow{1}{*}{25}  & 5.5 & \textbf{3.5} & 4.7 & \textbf{3.6}   \\ \hline
\multirow{1}{*}{40} & 12.0 & \textbf{8.1} & 11.2 & \textbf{8.4}   \\ \hline
\end{tabular}
\end{adjustbox}
\captionof{table}{Fine-tuning using instances from the train set with highest overlap with the test instances in DSADS dataset.}
\label{tab:overlapping}
\end{center}
\end{minipage}

\section{Conclusion and Future work}
In this work, we have studied the problem of adapting neural networks to varying input dimensionality in context of multivariate time series. 
This problem is of potential interest in several applications of deep learning in multivariate time series modeling where, despite capturing the behavior of the same underlying dynamical system, different instances of time series capture the behavior via a different set of sensors.
We map this problem to that of adapting the behavior of a core dynamical model of the underlying system via sensor-combination-specific conditioning vectors. The conditioning vector summarizes the available sensor combination for any time series via graph neural networks which inherently allow for combinatorial generalization.
Our results on three publicly available datasets prove the efficacy of the proposed approach to adapt to new sensor combinations.
Though we have evaluated the proposed approach for activity recognition and remaining useful life estimation applications, the proposed approach is generic, and can be useful in other multivariate time series applications where one or more dimensions may be missing at random across time series. Furthermore, it will be interesting to combine the proposed approach with other recent approaches that leverage the (known) underlying structure of the dynamical system being modeled, e.g. in \cite{narwariya2020graph}.

{\footnotesize
\bibliographystyle{named}
\bibliography{BibTex/sensor_analytics,BibTex/ijcai20,BibTex/aaai,BibTex/dataset}}

\end{document}